\documentclass[10pt,conference]{IEEEtran} 

\usepackage[utf8]{inputenc} 
\usepackage[T1]{fontenc}    
\usepackage{hyperref}       
\usepackage{url}            
\usepackage{booktabs}       
\usepackage{nicefrac}       
\usepackage{microtype}      

\usepackage{graphicx}
\usepackage{amsmath}
\usepackage{todonotes}
\usepackage{bm}
\usepackage{float}
\usepackage{subcaption}
\usepackage{textcomp}
\usepackage[numbers]{natbib}
\newcommand{\norm}[1]{\left\lVert#1\right\rVert}

\title{An Estimator for the Sensitivity to Perturbations of Deep Neural Networks}

\author{\IEEEauthorblockN{Naman Maheshwari\IEEEauthorrefmark{1}, 
Nicholas Malaya\IEEEauthorrefmark{2}, 
Scott Moe\IEEEauthorrefmark{2}, Jaydeep P. Kulkarni\IEEEauthorrefmark{1} 
and Sudhanva Gurumurthi\IEEEauthorrefmark{2}}
\IEEEauthorblockA{\IEEEauthorrefmark{1}Dept. of Electrical and Computer Engineering\\
University of Texas at Austin, Austin, TX 78712\\
Email(s): naman@utexas.edu, jaydeep@austin.utexas.edu}
\IEEEauthorblockA{\IEEEauthorrefmark{2}
  AMD Research \\
  Advanced Micro Devices, Inc. \\
  7171 Southwest Pkwy, Austin, TX 78735 \\
Email(s): nicholas.malaya@amd.com, scott.moe@amd.com, sudhanva.gurumurthi@amd.com}
}

\begin{document}

\maketitle

\begin{abstract}
  For Deep Neural Networks (DNNs) to become useful in safety-critical applications, such as self-driving cars and disease diagnosis, they must be stable to perturbations in input and model parameters. Characterizing the sensitivity of a DNN to perturbations is necessary to determine minimal bit-width precision that may be used to safely represent the network. However, no general result exists that is capable of predicting the sensitivity of a given DNN to round-off error, noise, or other perturbations in input. This paper derives an estimator that can predict such quantities. The estimator is derived via inequalities and matrix norms, and the resulting quantity is roughly analogous to a condition number for the entire neural network. An approximation of the estimator is tested on two Convolutional Neural Networks, AlexNet and VGG-19, using the ImageNet dataset. For each of these networks, the tightness of the estimator is explored via random perturbations and adversarial attacks.

\end{abstract}

\section{Introduction}
\label{sec:intro}
Deep neural networks (DNNs) achieve high accuracy in many machine learning tasks such as object recognition, speech recognition, and natural language processing. Many kernels in deep learning, particularly in convolutional neural networks (CNNs) are designed for computer vision tasks and are dominated by computation. For example, AlexNet \citep{KrizhevskyNIPS2012} has nearly 62.5 million parameters, 0.65 million neurons,  2.3 million weights (4.6 MB of storage), and requires 666 million MACs per 227x227 image (13 kMACs/pixel); whereas VGG-16 \citep{SimonyanICLR2015} possesses 14.7 million weights (29.4 MB of storage) and requires 15.3 billion MACs per 224x224 image (306kMACs/pixel) \citep{ChenISSCC2016}. The trend in current architectures is towards networks with more layers, which require more storage and computation per pixel. 

Lowering bit-widths ("Quantization") in both training and inference is therefore advantageous because it permits faster computation and reduced power use, particularly when the underlying hardware can natively support reduced precision.
However, higher performance must be balanced by the need for accuracy, particularly in safety-critical systems such as autonomous driving and health-care solutions where failure can be catastrophic. While many CNN architectures are resilient to lower precision, no rigorous and general result exists that provides \emph{a-priori} estimates of the accuracy impact from reduced precision. Similarly, inference is known to be more amenable to reduced precision than training, but no theoretical result exists explaining this observation \citep{HanICLR2016, RastegariECCV2016}.

Noting that modern DNNs are governed by computational arithmetic, and so are amenable to the tools of numerical analysis, the approach presented in this paper develops an estimator that predicts the impact of small perturbations to the inputs of a neural network on its output. This analysis must be performed on a network-by-network basis. This estimator can further be used to predict the minimal input precision required for that particular neural network such that the network does not become less stable from a perturbation to its input due to quantization. 


The paper is organized as follows. Related work in reduced precision analysis for neural networks is discussed in Subsection \ref{subsec:related}. Section \ref{sec:stability} derives an estimator of the condition number for a single neuron as a simple example, and then the analysis is expanded to a multi-layer neural network. Section \ref{sec:testing} discusses the techniques used to test the estimator with the aid of adversarial perturbation generation methods. Section \ref{sec:results} presents the estimator results applied to  two canonical CNN architectures, AlexNet and VGG-19 using both random perturbations and adversarial attack techniques to generate sample perturbations. Section \ref{sec:conc} concludes the work.

\subsection{Related work}
\label{subsec:related}

Recent research efforts have shown that neural networks are amenable to reduced precision. Prior DNN precision research work can be broadly classified into four categories. In the first, quantization techniques are explored for different neural network parameters during training and inference to compress the network with minimal loss of accuracy. The second is an empirical exploration of quantization for neural network parameters across different layers to achieve the same accuracy as the full precision model. The third is based on the derivation of theoretical bounds for neural networks with limited precision. The last category of prior work, which most closely relates to this paper, uses of the tools of numerical linear algebra such as matrix norms to study the properties of DNNs. The following four paragraphs summarize the related works in each of the four categories respectively.

(1) Training or inference with binary or ternary weights and activations can achieve comparable accuracy to full precision networks \citep{CourbariauxNIPS2015, LiArXiv2016, CourbariauxArXiv2016, RastegariECCV2016}. \citep{ZhouArXiv2016} tabulated various observations of different precision settings for weights and activations and also demonstrated how low precision can be used for gradients at training time. Recently, \citep{ChoiSysML2019} proposed 2-bit Quantized Neural Networks (QNNs) using techniques to individually target weight and activation quantizations which achieve higher accuracy than previous quantization schemes. \citep{OttArXiv2017} explored reducing the numerical precision of weights and biases for different recurrent neural networks (RNNs) empirically and concluded that weight binarization techniques are of limited use for RNNs while ternarization schemes yield similar accuracy to the baseline versions. 
All these methods help compress the size of the neural networks; however, they are unable to predict the impact of quantization \emph{a-priori}.
 
(2) \citep{JuddArXiv2015} studied per layer quantization and observed that the tolerance of CNNs to reduced precision data varies not only across networks, but also within layers. They proposed an empirical method to find a low precision configuration for a network while maintaining high accuracy. However, this method requires simulations and does not guarantee maintaining accuracy. \citep{JuddICS2016} is an extension of the previous work, where the authors proposed a method called Proteus which analyzes a given DNN implementation and maintains the native precision of the compute engine by converting to and from a fixed-point reduced precision format used in memory. This enables using different representation per layer for neuron activations and weights.  \citep{LaceyArXiv2018} presented a learning scheme to enable heterogeneous allocation of precision across layers for a fixed precision budget. The scheme is based on stochastic exploration for the DNN to determine an optimal precision configuration and also leads to  favorable regularization. However, the optimal value of the precision budget is again not known beforehand.

(3) \citep{SakrICML2017} derived theoretical bounds on the misclassification rate in the presence of limited precision. This work establishes bounds that limit misclassification after quantizing activations and weights to a fixed-point format from floating point. However, this work only bounds the accuracy loss between floating-point and fixed-point. \citep{GuptaICML2015} studied the impact of limited numerical precision on neural network training and the impact of rounding scheme in determining network’s behavior during training. It shows that 16-bit fixed-point representation incurs little accuracy degradation by using stochastic rounding but does not study the precision requirements during inference. \citep{ZhangICML2017} explored training at reduced precision but is mainly limited to linear models.

(4) This paper leverages the tools of numerical linear algebra, particularly stability analysis, to establish general bounds that can be imposed on the precision. Such an approach is not without precedent, and investigations of the properties imposed on a network by the measure of the weight matrix can be at least traced back to \citep{BartlettNIPS1996}, who showed that the generalization performance of a well-trained neural network with small training error depends on the magnitude of the weights, not the number. Reasoning about the generalization ability of a neural network in terms of the size, or norm, of its weight vector is called norm-based capacity control and \citep{NeyshaburCOLT2015} evaluated this for feed-forward neural networks. Along with capacity, they also investigated the convexity and characterization of the neural networks. \citep{BartlettNIPS2017} evaluated the spectral complexity of the neural networks using the Lipschitz constant (i.e., the product of the spectral norms of their weight matrices). \citep{LiangArXiv2017} showed that the Fisher-Rao norm provides an estimate of the size of the network weights and associates this with the trained network’s generalization capacity, which may be related to a network's susceptibility to perturbation or adversarial attack. Another recent work, \citep{LinICLR2019}, shows that error amplifications are a mode by which quantized models are prone to adversarial attacks. This empirical study also proposes a Defensive Quantization (DQ) method which controls the Lipschitz constant of the network during quantization. In contrast, this paper uses the measure of the weight matrix to study the stability of the neural networks to reduced precision and derive the precision requirements based on the estimator presented in Section \ref{sec:stability}.

\section{Stability Bounds on Forward Propagation}
\label{sec:stability}

In numerical analysis, the condition number of a system is a measure of the change in the output value of a function or a network for a small change in the input argument. If the condition number of a system is $\kappa(A) = 10^k$, then up to k digits of accuracy may be lost on top of the loss of precision from arithmetic methods. This work derives an analogue of the condition number for neural networks as,
\begin{equation}
    \label{eq:condition}
    \kappa = \bigg(\frac{\norm{\delta_y}/\norm{y}}{\norm{\delta_x}/\norm{x}}\bigg).
\end{equation}

Where $x$ and $y$ are the input and output to the network, respectively, and $\delta_x$ and $\delta_y$ are small perturbations to these quantities. The resulting quantity $\kappa$ estimates the susceptibility of a network to perturbations. 
We now derive $\kappa$ for a single neuron to demonstrate the methodology and generate intuition, and then generalize to an n-layer network. 

\subsection{Single Neuron Estimator Derivation}
\label{subsec:single}

Consider first a single neuron, neglecting bias. Each input ($x_1, x_2,\dots, x_n$) is multiplied by an associated weight ($\theta_1,\theta_2,\dots,\theta_n$). The results are then passed through an activation function, $f$ to produce the output, $y$.
This is expressed as,

\begin{equation*}
    y = f(\sum_i {\theta_i x_i}) = f(\theta_1x_1 + \theta_2x_2 + \dots + \theta_nx_n).
\end{equation*}

Consider a common and representative activation function, the rectified linear unit (ReLU),\footnote{The results shown subsequently apply to other common activation functions, and are detailed in Appendix \ref{appendices:appA}.}

\begin{equation*}
    f(z) = 
    \begin{cases}
        z, & \text{for } z > 0 \\
        0, & \text{for } z \leq 0 \\
    \end{cases}
\end{equation*}

which, due to its simplicity, has the added appeal of making subsequent computations more straightforward. Notice that ReLU is simply $\text{max}(0,z)$.

In the rest of this paper, we will use the shorthand $\theta \, x$ to indicate multiplication of a matrix of weights $\theta$ by a matrix $x$. Additionally, if $f$ is a scalar defined function and $x$ is a matrix, we will use the shorthand $f(x)$ to indicate $f$ applied to each entry of $x$. Then, the output of a single layer neural net satisfies,

\begin{equation}
    \label{eq:eq2}
    \norm{y} = \norm{f(\theta x)} = \norm{\text{max}(0,\theta x)}
\end{equation}

where $\norm{}$ denotes the generalized norm. To analyze the stability of the single layer neural network, we introduce a small perturbation, $\delta_x$, to the inputs. The resulting perturbation to the outputs is defined as $\delta_y$. Then,

\begin{equation*}
    \delta_y := f(\theta(x + \delta_x)) - f(\theta x),
\end{equation*}
\begin{multline}
    \label{eq:eq3}
    \norm{\delta_y} = \norm{f(\theta(x + \delta_x)) - f(\theta x)} \\
    = \norm{f(\theta x + \theta (\delta_x)) - f(\theta x)}.
\end{multline}

By the triangle inequality, 
\begin{multline}
    \label{eq:eq4}
    \norm{f(\theta x + \theta (\delta_x)) - f(\theta x)} \leq \\
    \norm{f(\theta x + \theta (\delta_x) - \theta x)} = \norm{f(\theta(\delta_x))}.  
\end{multline}

Using Equations \eqref{eq:eq2}, \eqref{eq:eq3} and \eqref{eq:eq4},

\begin{equation}
    \label{eq:eq5}
    \norm{\delta_y} \leq \norm{f(\theta(\delta_x)} \leq \norm{\theta\delta_x} \leq \norm{\theta}\norm{\delta_x},
\end{equation}
\begin{equation*}
    \frac{\norm{\delta_y}}{\norm{\delta_x}} \leq \norm{\theta}.
\end{equation*}

This result indicates that the amplification of a perturbation to the input of a single layer neural network is bounded by the norm of the weight matrix. However, this quantity is only significant relative to the overall magnitude of the data. For example, what if the network $f(\theta x)$ shrinks the magnitude of every input? Then, even though the quantity $\frac{\norm{\theta \delta y}}{\norm{\delta x}}$ is small, it may be a large perturbation relative to the magnitude of the initial quantity, $\frac{\norm{y}}{\norm{x}}$. This intuition drives the motivation that the pertinent quantity to be studied is therefore,

\begin{equation*}
	\frac{\norm{\delta_y}}{\norm{\delta_x}}/\bigg(\frac{\norm{y}}{\norm{x}}\bigg) = \bigg(\frac{\norm{\delta_y}/\norm{y}}{\norm{\delta_x}/\norm{x}}\bigg).
\end{equation*}

This is the quantity that was introduced in Equation \eqref{eq:condition}. Our estimator is the maximal value of this quantity, 

\begin{equation}
    \label{eq:estimator}
	\tilde{\kappa} = \max_{x \ne 0} \kappa = \max_{ x \ne 0}\bigg(\frac{\norm{\delta_y}/\norm{y}}{\norm{\delta_x}/\norm{x}}\bigg).
\end{equation}

Unfortunately, this estimator cannot be computed exactly unless it is known that $\norm{y} \ge C > 0,  \forall \, x\ne 0$ for some positive constant $C$.
Therefore, the estimator $\tilde{\kappa}$ is approximated by calculating a set of $\kappa$'s for many perturbations $\delta_x$ and inputs $x$. 

\subsection{Multi-layer Network Estimator Derivation}
\label{subsec:multi}

The analysis of the previous section is now generalized to many-layer networks. Stability and rounding error concerns become  more complicated when considering additional layers because each layer could introduce an error from rounding, which is amplified  in the subsequent layers. The multi-layer case has the form,

\begin{equation}
    \label{eq:eq7}
    \frac{\norm{\delta_y}}{\norm{\delta_x}} \leq \prod_{j=0}^{i-1} \norm{\theta_{n-j}}.
\end{equation}
Where $\norm{\theta_i}$ is the norm of the weight matrix of layer $i$ from the input side. Considering perturbations at every single layer due to rounding error for a simple feedforward neural network with n-layers  results in,
\begin{equation}
    \frac{\norm{\delta_y}}{\norm{\delta_x}} \leq \sum_{i=1}^n \left(\prod_{j=0}^{i-1} \norm{\theta_{n-j}}\right).
    \label{eq:multi-layer}
\end{equation}

\begin{figure}[ht]
    \centering
    \includegraphics[width=0.6\linewidth]{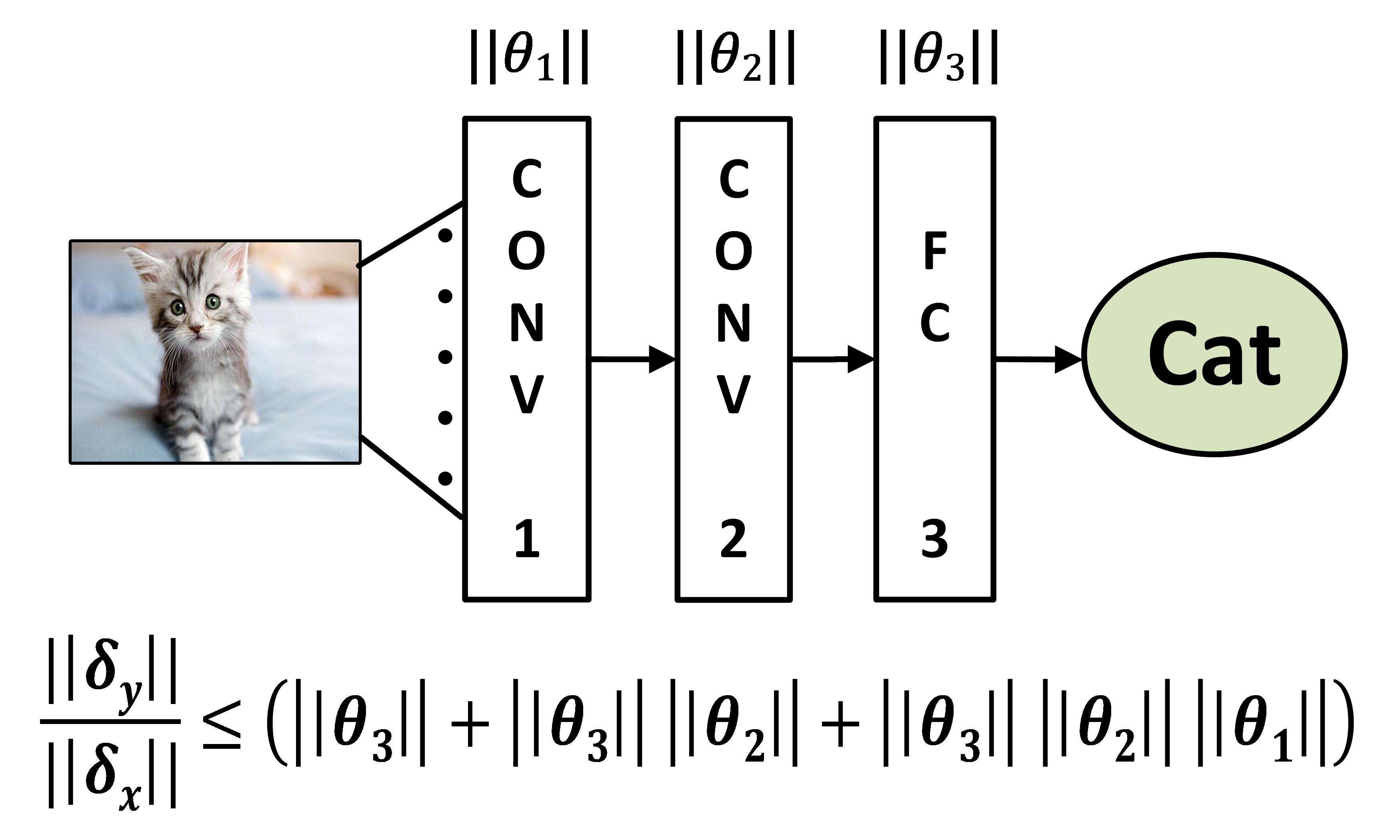}
    \caption{Example calculation of estimator in Equation \eqref{eq:multi-layer} for a 3-layer neural network. The first layer contributes to only one term, while the second layer contributes to two terms, and the last layer is a product of all the previous terms of the estimator.}
    \label{fig:KappaDemo}
\end{figure}

This is the product of all current and previous weight matrices, summed over each layer. The above equation indicates that the deeper layers cause the perturbations to grow because they amplify the rounding errors from previous layers. To aid in intuition, Figure \ref{fig:KappaDemo} shows the computation of $\frac{\norm{\delta_y}}{\norm{\delta_x}}$ for a simple three layer neural network.

It is interesting to note that many special layers, such as dropout, batchnorm, skip connections, etc., have the effect of reducing the magnitude of the weights. These layers may act as a mechanism to reduce the condition number of the overall network, and in doing so, render the overall network more stable and less susceptible to perturbations.

\section{Testing the Estimator with Adversarial Perturbations}
\label{sec:testing}

The estimator presented in Equation \eqref{eq:multi-layer} provides an upper-bound of the susceptibility of a network to a perturbation in input. A natural question is: how tight is this bound to perturbations in the network? If the estimator is not tight, it will overestimate the impact of perturbations. To explore the tightness of the estimator presented previously, we seek to minimize the quantity, $\norm{\delta_x}$. In particular, the smallest perturbations $\norm{\delta_x}$ that in turn maximizes the quantity $\norm{\delta_y}$. To efficiently generate such perturbations, we leverage the existing body of work on adversarial perturbations. We next discuss the techniques we use to generate these perturbations in Subsection \ref{subsec:adversarial}, and then present the results of these perturbations in Subsection \ref{subsec:generated}. 

\subsection{Techniques to Generate Adversarial Perturbations}
\label{subsec:adversarial}

Adversarial attacks are methods designed to alter the solution or classifier output of a learning system. For CNNs, an adversarial perturbation, $\Delta(\bm{x}; \hat{k})$, refer to small perturbations $\bm{r}$ added to an input image $\bm{x}$ such that the network classifier $\hat{k}(\bm{x})$ changes, leading to misprediction. This is formally presented in Equation \eqref{eq:eq9} as,

\begin{equation}
    \label{eq:eq9}
    \Delta(\bm{x}; \hat{k}) = \underset{\bm{r}}{\text{min}}\norm{\bm{r}} \text{subject to } \hat{k}(\bm{x} + \bm{r}) \ne \hat{k}(\bm{x}). 
\end{equation}

Existing adversarial attacks are used as they represent the best-known methods to reliably produce misclassification from small perturbations in input. Existing techniques also have the advantage of prior peer-review, making the results more accepted and broadly accessible to the community. Based on these criteria, this work focused on the common adversarial attack, DeepFool, which we discuss below.

\citep{DezfooliCVPR2016} proposed the untargeted attack technique known as DeepFool. This method is based on an iterative linearization of the classifier to generate minimal perturbations sufficient to change the classification label. Initially, it is assumed that the neural networks are completely linear, with classifiers separated by  hyperplanes. Since neural networks are non-linear, the linearization process is iterated until the classification index changes. In this work, the iterator was observed to converge in less than four iterations for most images. The process of generating the perturbations is computationally inexpensive and this is therefore an effective technique to generate small adversarial perturbations. 

\citep{DezfooliCVPR2017} presents an extension of DeepFool which generates a small, image-agnostic perturbation vector which causes misclassification on a large set of images across a wide variety of classifiers. This means that an image-specific perturbation vector need not be generated, and a universal perturbation when added to different input images can cause misclassification with probability of about 70\% across different networks. This is particularly interesting from the perspective of this paper, because our method directly provides a bound on the magnitude of the change in the output of the network caused by the universal perturbation. In this way, we expect our estimator to apply to a wide range of classifiers impacted by the universal perturbation generated by DeepFool.

\subsection{Generated DeepFool Perturbations}
\label{subsec:generated}

This analysis used pre-trained models of two standard CNNs designed for object recognition tasks on the ILSVRC2012 ImageNet dataset \citep{ILSVRC15}:  AlexNet and VGG-19. These networks and the associated weights were taken from the BVLC Caffe2 models publicly available via ONNX \citep{ONNX}. All the analysis was performed on an AMD Radeon Pro Vega Frontier Edition.

\begin{figure}[ht]
    \centering
    \begin{subfigure}[ht]{0.48\linewidth}
    \centering
    \includegraphics[width=\linewidth]{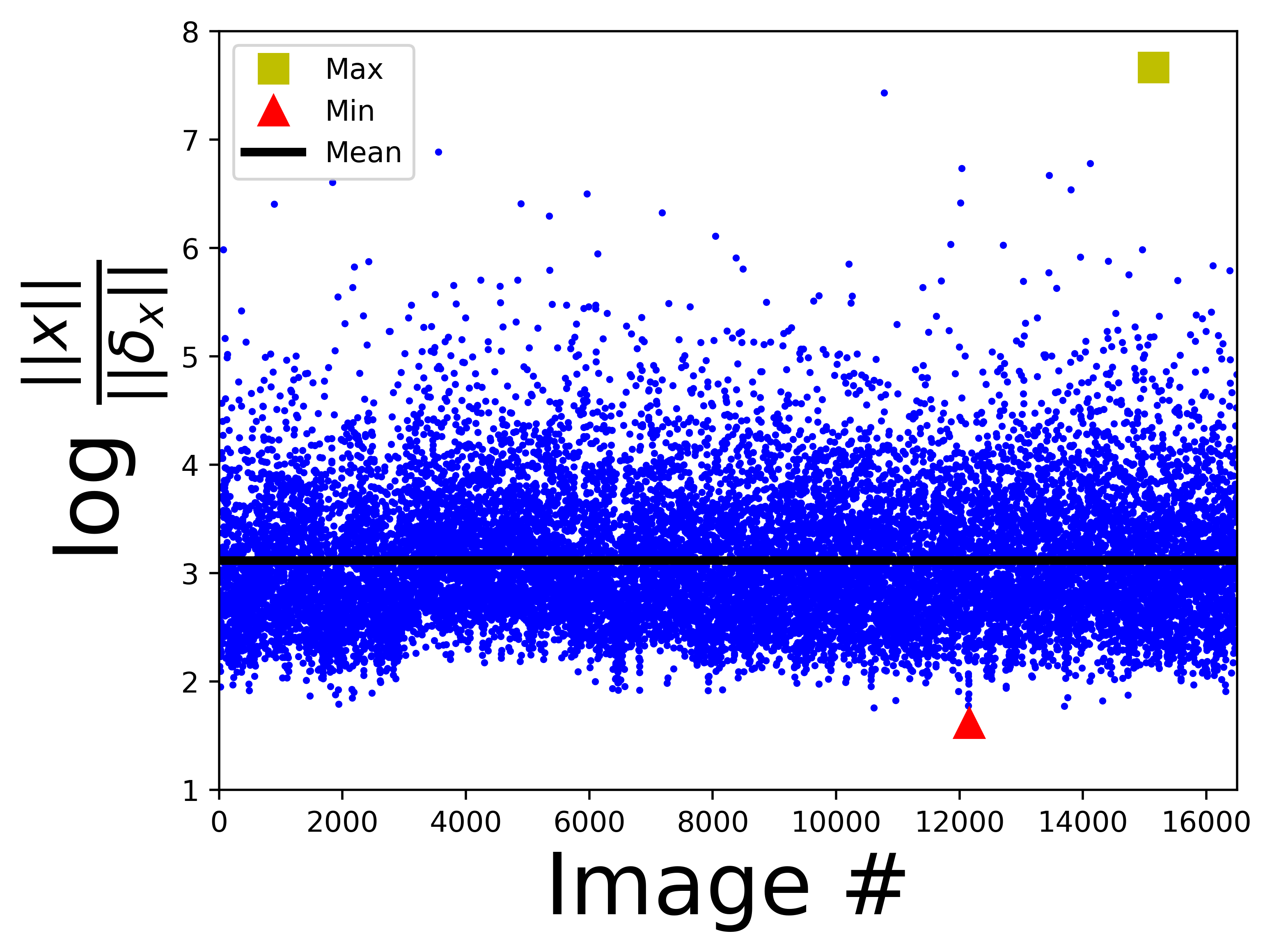}
    \caption{AlexNet}
    \label{fig:DeepFool_AlexNet}
    \end{subfigure}
~
    \begin{subfigure}[ht]{0.48\linewidth}
    \centering
    \includegraphics[width=\linewidth]{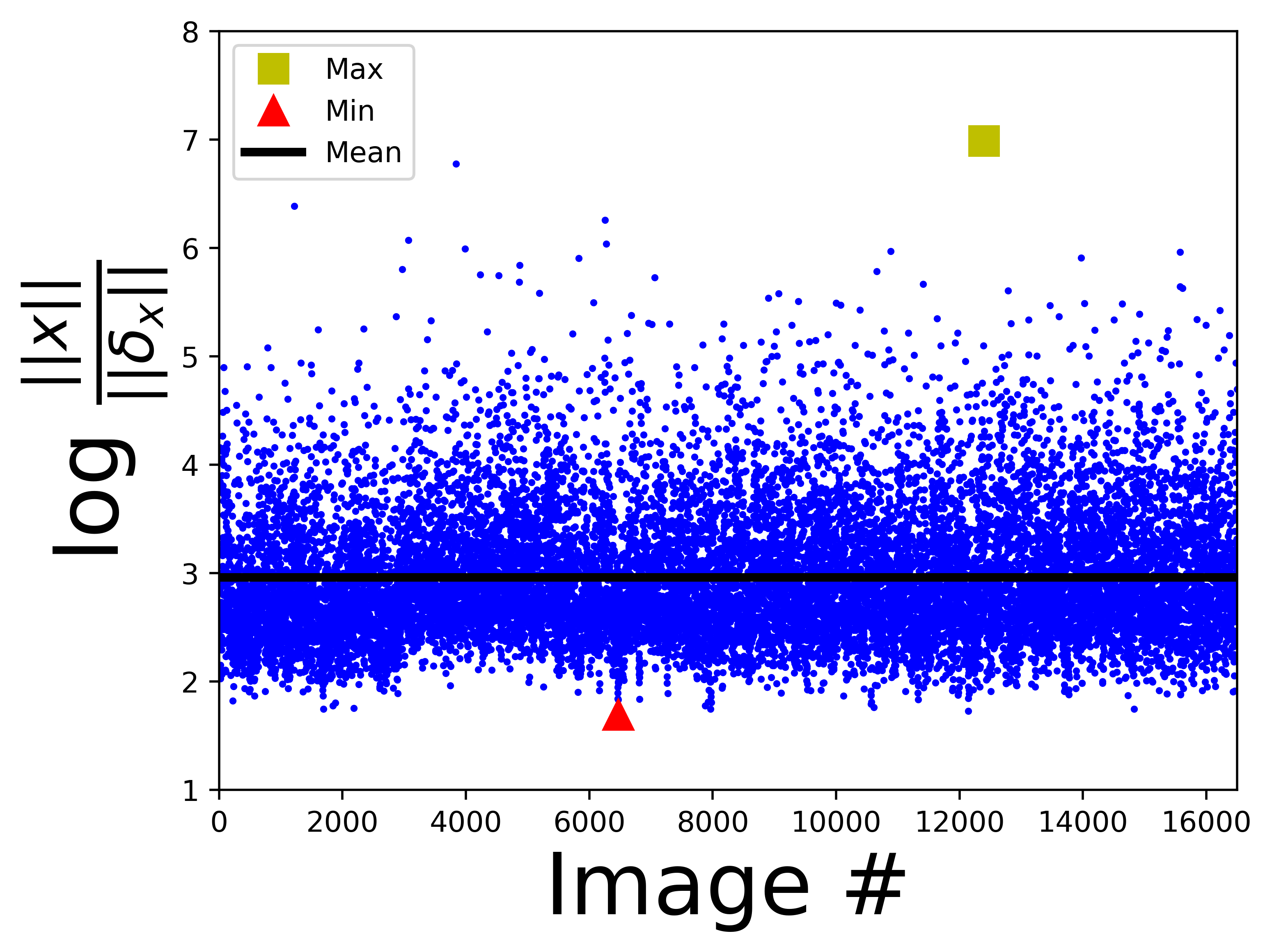}
    \caption{VGG-19}
    \label{fig:DeepFool_VGG19}
    \end{subfigure}
    
    \caption{Relative magnitudes of generated DeepFool perturbations that resulted in misclassification. These images indicate that the generated DeepFool perturbations $\delta_x$ are indeed significantly smaller than the magnitude of the original image, $x$. Notice that the perturbation sizes range several orders of magnitude.}
    \label{fig:DeepFoolSizes}
\end{figure}

Figure \ref{fig:DeepFoolSizes} shows the magnitudes of DeepFool perturbations generated for AlexNet and VGG-19 measured relative to the original image, $\norm{\delta_x}$, over the norm of the input image, $\norm{x}$. Since this quantity is much less than 1, the results are plotted by the reciprocal on a logarithmic scale, or $\log(\frac{\norm{x}}{\norm{\delta_x}})$. The x-axis spans different images, and the y-axis details the resulting perturbation size.  For this analysis, 16,500 test images were used. 
This sampled ImageNet with 20 randomly chosen images from each of the randomly chosen 825 classes out of 1000. The ordering on these plots is such that 20 images from one class are represented adjacent to each other. Notice that the data points corresponding to the generated perturbations for images of a given class are closely located, showing that there exists correlations in the data. This is not unexpected, as images from a common classifier share characteristics that also impact $\norm{x}$, such as common pixel color. In turn, this implies that it is easier to generate minimal adversarial perturbations for some classes than others, and that images of the same classifier have similar sensitivities to perturbations.

The smallest perturbations relative to the image are of the order of $10^{-8}$ for AlexNet and $10^{-7}$ for VGG-19. On average, the perturbations are of the order of $10^{-3}$ for the two networks. 
Note that all these generated perturbations, when added to the original image, lead to misclassification by the network. Anecdotally, a few of the resulting perturbed images were visually inspected by humans, and did not have any noticeable artifacts or distortions.
\section{Results}
\label{sec:results}
This section compares the condition number (calculated with Equation \eqref{eq:estimator}) to the generalized small perturbation vectors from the previous section. Note that the quantities $\norm{\delta_y}/\norm{\delta_x}$ and  $\norm{\delta_y}$ are computed as the magnitudes of the element-wise difference of the pre-final output layer (i.e. the layer before the softmax layer between the original and perturbed scenarios). We then compare the estimator to empirical data generated via adversarial attacks and random noise.
\subsection{DeepFool perturbations}
\label{subsec:res_df}
\begin{figure}[H]
    \centering
    \begin{subfigure}[ht]{0.48\linewidth}
    \centering
    \includegraphics[width=\linewidth]{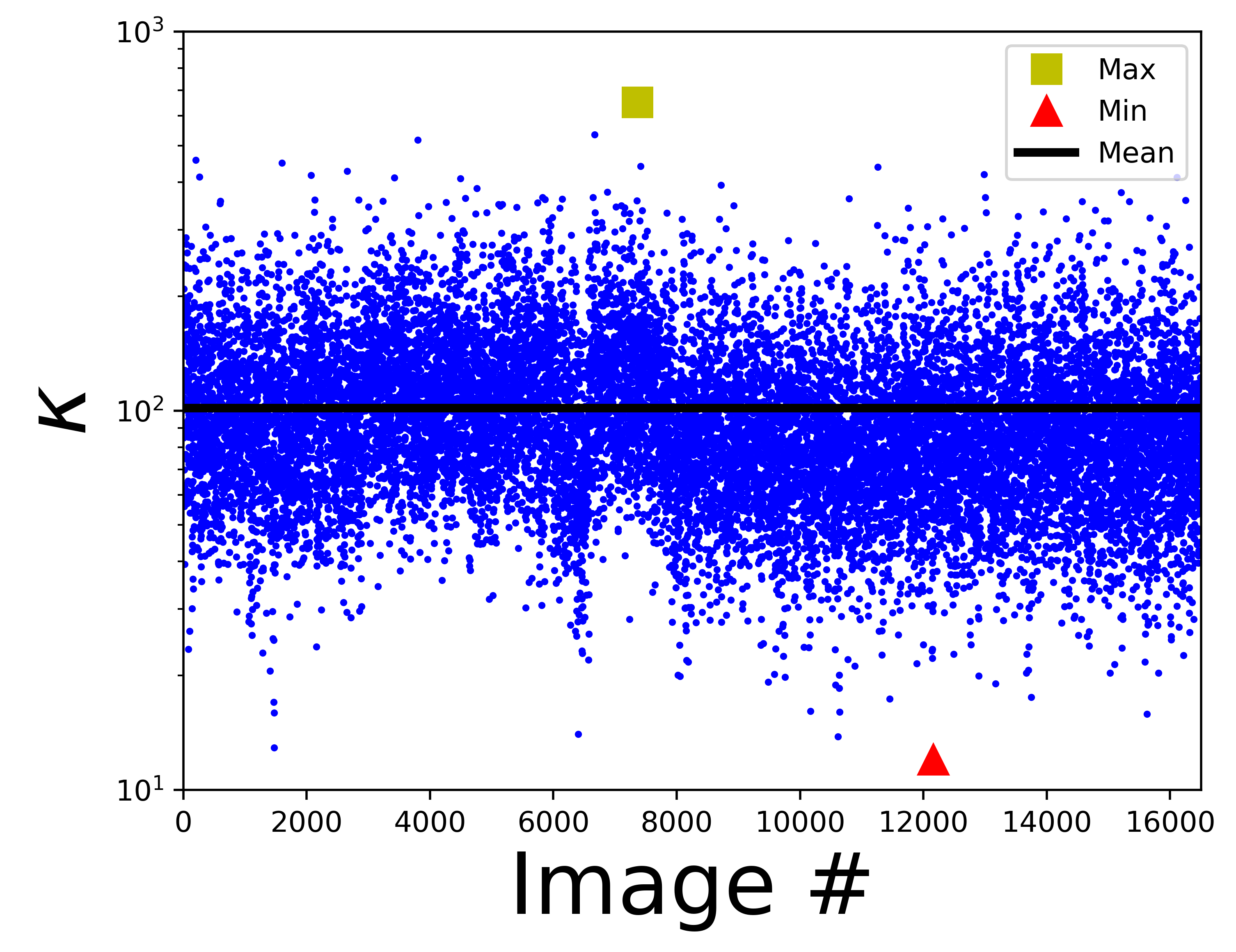}
    \caption{AlexNet}
    \label{fig:Estimator_AlexNet}
    \end{subfigure}
~
    \begin{subfigure}[ht]{0.48\linewidth}
    \centering
    \includegraphics[width=\linewidth]{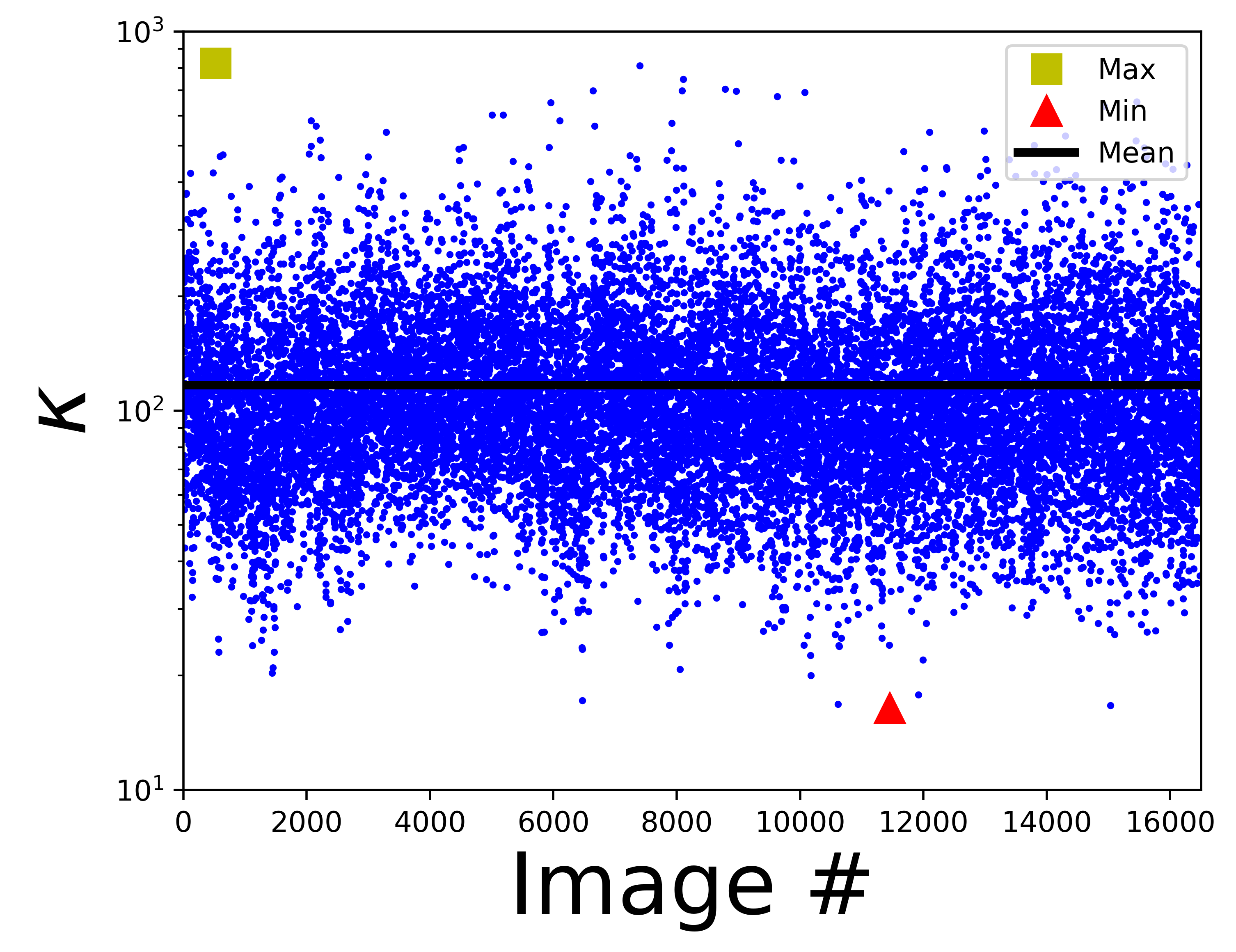}
    \caption{VGG-19}
    \label{fig:Estimator_VGG19}
    \end{subfigure}
    
    \caption{$\kappa$ estimator values. The X-axis is the image number and Y-axis shows the the maximum value of $\kappa$ created using the adversarial perturbations discussed in Section \ref{subsec:generated} for each image. All of the 16,500 data are rendered as blue circles. The red triangle indicates the perturbation that resulted in the smallest change in output to the network. The yellow square indicates the perturbation that caused the largest change in the network output. These plots are similar, indicating that the networks have similar susceptibility to perturbations in input. For either network, the range of results spans many orders of magnitude, indicating a wide range of possible impact from a perturbation.}
    \label{fig:Estimator}
\end{figure}

The quantity $\kappa$ presented in Equation \eqref{eq:estimator} was computed for AlexNet and VGG-19 using the generated DeepFool perturbations applied to each of the 16500 images discussed in Section \ref{subsec:generated}. The $\kappa$ values are shown in Figure \ref{fig:Estimator}. The maximum $\kappa$ value created by applying the DeepFool perturbations to inputs to AlexNet is about 651, the average image has a maximum $\kappa$ value of about 100, and all images have a maximum $\kappa$ value of at least of 12.
For VGG-19, the maximum $\kappa$ value is about 825, the average image has a maximum $\kappa$ value of about 117, and all images have a maximum $\kappa$ value of at least 16.

Table \ref{table:table2} shows the precision requirements for input representations of AlexNet and VGG-19 based on the  mean, maximum and minimum value of $\kappa$. The maximum value of $\kappa$ gives maximum amount a small perturbation to the input of the neural network amplified by the network. 

In practice, the error in the output of the network can be expected to be $\epsilon \tilde{\kappa}$, where $\epsilon$ is machine epsilon and $\tilde{\kappa}$ is defined in Equation \eqref{eq:estimator}. Thus, $\tilde{\kappa}$ gives an estimate of the minimum precision that should be used with a particular neural network. The minimum digits required can be calculated as $log_{10}(\kappa)$. The minimum number of bits is calculated as, $\lceil log_2(\kappa) \rceil + 1$.
\begin{table}[ht]
  \caption{Precision requirements for input representations based on DeepFool perturbations.}
  \label{table:table2}
  \centering
  \begin{tabular}{ccc}
    \toprule
    \cmidrule(r){1-2}
    \textbf{Network} & \textbf{AlexNet} & \textbf{VGG-19} \\
    \midrule
    \textbf{Mean $\kappa$} & 101.78 & 116.84 \\
    \textbf{Minimum Digits}  & 2.01 & 2.07 \\
    \textbf{Minimum Bits}  & 8 & 8 \\
    \midrule
    \textbf{Maximum $\kappa$} & 651.06 & 824.53 \\
    \textbf{Minimum Digits}  & 2.81 & 2.92 \\
    \textbf{Minimum Bits}  & 11 & 11 \\
    \midrule
    \textbf{Minimum $\kappa$} & 12.00 & 16.37 \\
    \textbf{Minimum Digits}  & 1.08 & 1.21 \\
    \textbf{Minimum Bits}  & 5 & 6 \\    
    \bottomrule
  \end{tabular}
\end{table}

For the worst-case generated perturbations, ten bits of precision could be lost for both AlexNet and VGG-19 and hence, 11 bits are required for input representations at a minimum. For the mean scenario, eight bits of precision are required for both networks, whereas the minimum $\kappa$ says that AlexNet requires five bits of precision while VGG-19 requires six bits. This shows that moving to \textit{"INT8"} (eight-byte integer precision), a commonly supported precision in most processors, for representing inputs to these networks may make them more likely to misclassify inputs. 

\subsection{Random perturbations}
\label{subsec:res_random}

The DeepFool perturbations are carefully constructed to cause misclassification by the network. It is expected that more indiscriminate sources of perturbations, such as rounding error or noisy data, will have lesser impact on the output of neural networks. To test this, 16,500 perturbations of same magnitude as the DeepFool perturbations, but with random directions,  were generated. Those perturbations were added those to the same set of images used in Section \ref{subsec:res_df} and again the  maximum value of $\kappa$ was estimated for each image. In this case, the $\kappa$ values observed for these random perturbations are much smaller than from adversarial perturbations. For example, the mean $\kappa$ is reduced from approximately 101 to less than 2 in AlexNet, and from approximately 116 to 3. This suggests that for applications where the consequences of infrequent misclassification are not severe, lower precision than $11$ bits may suffice. However, networks more aggressively quantized in this manner are likely more susceptible to adversarial attacks \citep{LinICLR2019}.


\section{Conclusions and Future Work}
\label{sec:conc}


In this paper, an estimator was derived which can predict the sensitivity of a neural network to perturbations. These estimators can help in estimating the minimum precision requirements for input representations of various neural networks. We show the results for two widely studied CNN architectures, AlexNet and VGG-19, across both adversarial attacks and random perturbations. 
The estimator can be used to guide decisions of precision support required for hardware when designing deep learning accelerators, thus enabling energy-efficient edge computing where power consumption is a major bottleneck. At the software level, the estimator can be used to estimate the maximum evaluation error required for certification, validation, or quantification of uncertainty. At the algorithmic level, the estimator enables the design of efficient DNN architectures which can be resistant to noise or adversarial attacks.

Future work should consider introducing perturbations at each layer to mimic perturbing the weights. Further investigations should also be performed to extend the estimator to the precision requirements for activations, weights, and biases for each layer. Finally, the estimator should be extended to a wide variety of neural networks such as Recurrent Neural Networks (RNNs), Reinforcement Learning (RL), Generative Adversarial Networks (GANs), etc. to consider and characterize the stability of these architectures.

\subsubsection*{Acknowledgments}

AMD, the AMD Arrow logo, and combinations thereof are trademarks of Advanced Micro Devices, Inc. Other product names used in this publication are for identification purposes only and may be trademarks of their respective companies.

\textcopyright 2019 Advanced Micro Devices, Inc. All rights reserved.

\bibliographystyle{IEEEtran}
\bibliography{IEEEabrv,epsilon} 

\appendix
\section*{Derivation of bound for other activation functions}
\label{appendices:appA}

The bound presented in section \ref{subsec:single} was derived for the most common activation function, Rectified Linear Unit (ReLU). Here, we derive the bounds for other commonly used activation functions as well and show that the results presented in the paper hold true.

\subsection{Leaky Rectified Linear Unit (Leaky ReLU)}

\begin{equation*}
    f(z) = 
    \begin{cases}
        z, & \text{for } z > 0 \\
        \alpha z, & \text{for } z \leq 0; \text{ $\alpha$ = small const. (e.g., 0.1)} \\
    \end{cases}
\end{equation*}

\begin{equation*}
    \norm{f(z)} = \norm{\text{Max}(\alpha z, z)} \leq \norm{z}    
\end{equation*}

\subsection{Exponential Linear Unit}

\begin{equation*}
    f(z) = 
    \begin{cases}
        z, & \text{for } z > 0 \\
        \alpha (e^{z}-1), & \text{for } z \leq 0; \text{ $\alpha$ = small const. (e.g., 0.1)} \\
    \end{cases}
\end{equation*}

\begin{equation*}
    0 < \norm{e^{z}} \leq 1 \text{ for } z \leq 0
\end{equation*}
\begin{equation*}
        0 \leq \norm{e^{z}-1} < 1 \text{ for } z \leq 0
\end{equation*}
\begin{equation*}
    \norm{a}\norm{e^{z}-1} \leq \norm{\alpha} \text{ for } z \leq 0
\end{equation*}

\begin{equation*}
    \norm{f(z)} = \norm{\text{Max}(\alpha , z)} 
\end{equation*}

\subsection{Sigmoid}

\begin{equation*}
    f(z) = \frac{1}{1+e^{-z}}
\end{equation*}
\begin{equation*}
    \norm{1+e^{-z}} > 1
\end{equation*}
\begin{equation*}
    \norm{f(z)} < 1
\end{equation*}

\subsection{Hyberbolic Tangent}

\begin{equation*}
    f(z) = \frac{e^{z}-e^{-z}}{e^{z}+e^{-z}}
\end{equation*}
\begin{equation*}
    \norm{e^{z}} > 0, \norm{e^{-z}} > 0
\end{equation*}
\begin{equation*}
    (\norm{e^{z}} - \norm{e^{-z}}) < (\norm{e^{z}} + \norm{e^{-z}})
\end{equation*}
\begin{equation*}
    \norm{f(z)} < 1
\end{equation*}


\end{document}